\documentclass{sig-alternate-05-2015}
\usepackage{graphicx}
\usepackage{amsmath}
\usepackage{amsfonts}
\usepackage{comment}
\usepackage{multirow}
\usepackage{array}
\usepackage{subcaption}
\newcommand\numberthis{\addtocounter{equation}{1}\tag{\theequation}}

\begin{document}

\setcopyright{acmcopyright}




%

\title{Drug-Drug Interaction Extraction from Biomedical Text Using Long Short Term Memory Network}
%
%
%
%
%

\numberofauthors{8} 
%
\author{ 
\alignauthor
Sunil Kumar Sahu\\
       \affaddr{IIT Guwahati, India}\\        
       \email{sunil.sahu@iitg.ernet.in}
\alignauthor
Ashish Anand\\
       \affaddr{IIT Guwahati, India}\\       
       \email{anand.ashish@iitg.ernet.in}
 }

\maketitle
\begin{abstract}
Simultaneous administration of multiple drugs can have synergistic or antagonistic effects as one drug can affect activities of other drugs. Synergistic effects lead to improved therapeutic outcomes, whereas, antagonistic effects can be life-threatening, may lead to increased healthcare cost, or may even cause death. Thus identification of unknown drug-drug interaction (DDI) is an important concern for efficient and effective healthcare. Although multiple resources for DDI exist, they are often unable to keep pace with rich amount of information available in fast growing biomedical texts. Most existing methods model DDI extraction from text as a classification problem and mainly rely on handcrafted features. Some of these features further depend on domain specific tools. Recently neural network models using latent features have been shown to give similar or better performance than the other existing models dependent on handcrafted features. In this paper, we present three models namely, {\it B-LSTM}, {\it AB-LSTM} and {\it Joint AB-LSTM} based on long short-term memory (LSTM) network. All three models utilize word and position embedding as latent features and thus do not rely on explicit feature engineering. Further use of bidirectional long short-term memory (Bi-LSTM) networks allow implicit feature extraction from the whole sentence. The two models, {\it AB-LSTM} and {\it Joint AB-LSTM} also use attentive pooling in the output of Bi-LSTM layer to assign weights to features. Our experimental results on the SemEval-2013 DDI extraction dataset show that the {\it Joint AB-LSTM} model outperforms all the existing methods, including those relying on handcrafted features. The other two proposed LSTM models also perform competitively with state-of-the-art methods.
\end{abstract}

%
%

%
%

%
%
\printccsdesc



\section{Introduction}
There has been a significant rise in a number of persons taking multiple drugs at the same time. According to the numbers released in 2010 by the US Centers for Disease Control and Prevention, one in ten Americans is on five or more medications \cite{hutson11}. Similar statistics can be expected from other countries as well. When multiple drugs are administered together, there is an inevitable risk of a drug affecting activities of other drugs. Effect of DDI can be either synergistic or antagonistic. Adverse drug reaction (ADR) is an example of antagonistic effect~\cite{businaro2013we}. With the rise in people taking multiple drugs, it is very important to have DDI information available in structured form. DrugBank\footnote{\url{https://www.drugbank.ca/}} and Stockley\footnote{\url{https://www.medicinescomplete.com/mc/alerts/current/drug-interactions.htm}} are examples of knowledge bases (KBs) keeping DDI information in a structured form. However, keeping KBs update with growing rate of biomedical literature is a challenging task~\cite{segura2011,segura2013}. PubMed, a database of biomedical articles, contains about 27 million citations\footnote{\url{https://www.ncbi.nlm.nih.gov/pubmed}} and approximately $0.8$ million articles are being added annually\footnote{\url{https://www.nlm.nih.gov/bsd/stats/cit_added.html}}.  The growing rate of this number in recent years has necessitated the development of efficient automatic tools for information extraction across biomedical domain \cite{przybyla16,singhal16}.

Identifying DDIs in text is the process of recognizing how two drugs in a given sentence are related \cite{segura2013}. We illustrate different interaction types between two drugs through examples in Table\ref{tab:ddi_example}. The two pairs ({\it Fluoxetine}, {\it Phenelzine}) and ({\it Crocin}, {\it Phenelzine} ) in the sentence [$S_1$] fall into \textit{Advice} interaction category as they are suggested to be not taken together.
Similarly, interacting drug pairs ({\it PGF2alpha}, {\it Oxytocin}) in the sentence [$S_2$] and ({\it Ketamine}, {\it Halothane}) in [$S_3$] belong to {\it Effect} and {\it Mechanism} categories respectively, since impact and mechanism of impact are present in the respective sentences. Sentence [$S_4$] does not say anything more than that the two drugs ({\it Warfarin} and {\it Rifampin}) are interacting, so it falls into {\it Interaction}
 ({\it Int}) class. 
Identifying this kind of information can also be useful for other applications such as drug repurposing, semantic search and other information retrieval tasks.
 
\begin{table*}[hbt]
\begin{minipage}{\textwidth}
\centering
\scalebox{0.9}{
\begin{tabular}{|l|l|l|} 
[$S_1$] & {\bf Fluoxetine} and crocin should not be administered to patients receiving {\bf phenelzine} & {\it Advice(Fluoxetine, Phenelzine)}
\\ \cline{0-2}
[$S_1$] & Fluoxetine and {\bf crocin} should not be administered to patients receiving {\bf phenelzine} & {\it Advice(Crocin, Phenelzine)} \\ \cline{0-2}
[$S_2$] & Both {\bf PGF2alpha} and {\bf Oxytocin} induced dopamine release in the nucleus accumbens & {\it Effect(PGF2alpha, Oxytocin)} 
\\ \cline{0-2}
[$S_3$] & The half life of {\bf ketamine} in plasma and brain was longer in the presence of {\bf halothane}& {\it Mechanism(Ketamine, Halothane)}
\\  \cline{0-2}
[$S_4$] & The drug interaction between {\bf warfarin} and {\bf rifampin} is not well known & {\it Int (Warfarin, Rifampin)} 
\\
\end{tabular}
}
\caption{Examples illustrating different types of interactions between two drugs. All examples are taken from the dataset described in Sec.~\ref{dataset}.}
\label{tab:ddi_example}
\end{minipage}
\end{table*}

Realizing the importance of automatic extraction of drug interaction information, two challenges were organized. The first DDI extraction challenge~\cite{segura2011}, organized as a workshop in the SEPLN 2011\footnote{\url{http://www.uhu.es/sepln2011/index.php/en.html}} conference, focused on the DDI extraction task. The aim of the task was to identify whether there was an interaction between a given pair of drugs mentioned within a sentence. It was assumed that the drug names present in sentences were given. Extension of the first challenge was organized as SemEval 2013 Task-9~\cite{segura2013}. In this challenge, two tasks were designed: first task concentrated on drug name recognition and their classification, and the second focused on DDI classification from biomedical text. In this work, we focus on the second task of the SemEval 2013 Task-9 only.

Existing methods can be classified into two categories: one-stage and two-stage methods. In one-stage methods \cite{bobic2013,hailu2013,thomas2013}, a multi-class classifier is used to map a sentence with two target drugs either into one of the interacting classes or into the negative class. On the other hand two-stage methods~\cite{rastegar2013,bjorne2013}, as the name suggests, break the problem into two steps. The first step builds a binary classifier to determine whether an interaction exists between two target drugs or not. Only those sentences with target drug pairs, which fall into positive category of the binary classifier in the previous step, are considered as input to multi-class classifier of the second step. These methods can further be divided into two categories, methods relying on handcrafted features and methods using latent features. In the first category mainly support vector machines (SVMs) with linear or non-linear kernels have been used in several studies \cite{chowdhury2013,bokharaeian2013,kim2015}. All of these methods are dependent on manually engineered features such as PoS tag, chunk tag, trigger words, shortest dependency tree and syntax tree. Methods using non-linear kernels map structure features (dependency tree and syntax tree) into real values. Such methods have been successfully used for other similar relation extraction tasks including ADRs extraction from biomedical texts \cite{gurulingappa2013,gurulingappa2012,harpaz2014,xu2015} and from social media texts \cite{yang2015}, protein-protein interaction extraction from biomedical text~\cite{Qian2012}, relation between genes and diseases~\cite{bravo2015}, and relations between medical concepts~\cite{rink2011}.
Although such methods have been shown to perform well, they require manually crafted features. Extraction of these features is however dependent on other NLP tools. Inherent noise and cost of such tools may adversely affect the performance of models dependent on these features.  
Methods using latent features and belonging to the second category, are result of re-emergence of deep learning models as a powerful alternative to conventional feature based models. Some notable studies \cite{liu2016,zhao2016} for the DDI extraction tasks are based on convolution neural networks (CNNs) and have been shown to achieving superior performance than the existing state-of-the-art methods. We discuss more about these methods in section~\ref{sec:baseline}.

In this work, we also rely on latent features learned by neural network models. As opposed to works in \cite{liu2016,zhao2016}, which used CNN models, we use LSTM based neural network models \cite{Hochreiter97}. CNN models require pooling on continuous $n$ grams built on entire sentence to obtain constant length features. Here $n$ is the length of convolution or filter. It may cause problems for the sentences of large length and/or for sentences having important clues lying far away from each other. To overcome this issue we use Bi-LSTM with two different pooling techniques for encoding variable length features. Theoretically, a Bi-LSTM can preserve information about the past and future words while reading \cite{Hochreiter97}. Therefore when we apply pooling on the output of Bi-LSTM, we can get features containing information about complete context from whole sentence. This is in contrast to the CNN models which extract features based on $n$ gram of the sentence. With this intuition, we propose three models, namely: {\it B-LSTM}, {\it AB-LSTM} and {\it Joint AB-LSTM} for the DDI extraction task. Here {\it B-LSTM} and {\it AB-LSTM} uses a Bi-LSTM for encoding word and position features. {\it B-LSTM} uses max pooling and {\it AB-LSTM} uses attentive pooling on the outputs of Bi-LSTM to get fixed length features over complete sentence. On the other hand, {\it Joint AB-LSTM} being an ensemble of {\it B-LSTM} and {\it AB-LSTM} uses two Bi-LSTMs, one with max pooling and another with attentive pooling. In each of these models we use fully connected neural network in output layer.
 
All the three proposed models give either competitive performance compared to the existing methods or have achieved new state-of-the-art performance. The two important features of the models are: all of them belong to one-stage category and use simple features. None of the chosen features explicitly extract syntactic information hidden in a sentence. Among the three proposed models, {\it Joint AB-LSTM} outperforms all existing models for the DDI extraction tasks. Analysis of our results indicate that all models finds it difficult to make correct prediction for drug pairs present in a long sentence having too many other drug entries. If we compare between CNN and LSTM models, then LSTM models are generally found to have better prediction for longer sentences than the CNN model. We believe that new models should work in the direction of mitigating above issues to bring significant improvement. 

\section{Model Architecture}
\label{model}
\begin{figure*}[bt]
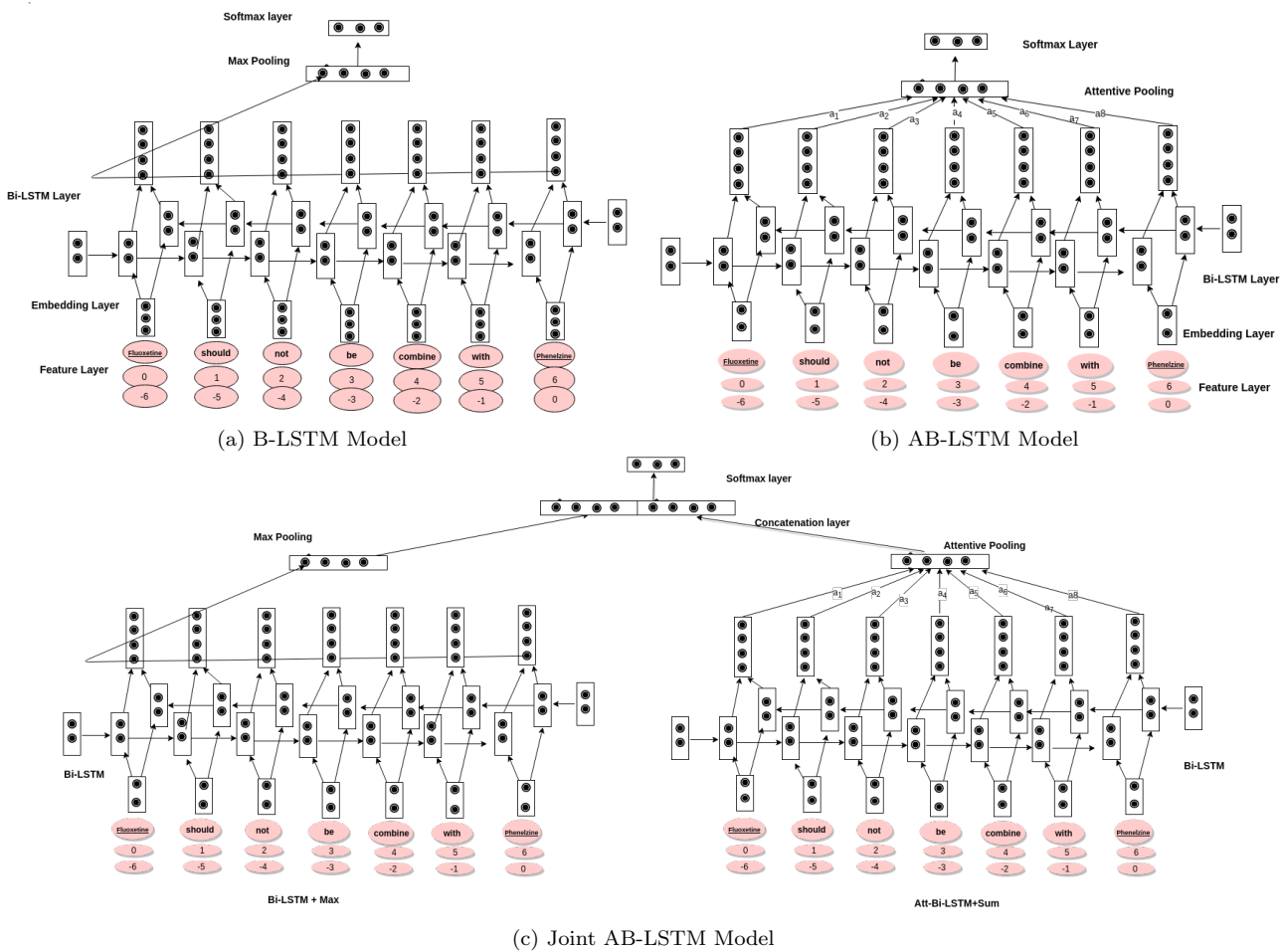

\centering
  \begin{subfigure}[b]{.5\linewidth}
    \centering
    \includegraphics[width=0.95\textwidth]{b_lstm.png}
    \caption{B-LSTM Model}\label{fig:1a}
  \end{subfigure}%
  \begin{subfigure}[b]{.5\linewidth}
    \centering
    \includegraphics[width=0.95\textwidth]{ab_lstm.png}
    \caption{AB-LSTM Model}\label{fig:1b}
  \end{subfigure}%
  \\%
  \begin{subfigure}[b]{1.0\linewidth}
    \centering
    \includegraphics[width=.9\textwidth]{joint_ab_lstm.png}
    \caption{Joint AB-LSTM Model }\label{fig:1m}
  \end{subfigure}%
  \caption{Block diagram of all three models }\label{fig:models}
\end{figure*}

We present three LSTM based models namely, {\it B-LSTM}, {\it AB-LSTM} and {\it Joint AB-LSTM} for the DDI extraction task. We assume that the two targeted
drug names are given in a sentence and model has to classify it into one of the five categories: {\it Advice},
{\it Effect}, {\it Mechanism}, {\it Int}, {\it Negative}. Section~\ref{dataset} describes about the task in detail. 
Architecture of the three proposed models are shown in Figure \ref{fig:models}. 
Each model uses embedding features as input in first layer and learn fixed length vector representation through subsequent layers. Score for each possible class is computed in the final layer and decision is made using these scores. We now briefly explain each components of the three models.  
\subsection*{Feature Layer}
We represent each word in the sentence with three discrete local features, namely: {\it word} (W), {\it distance$_1$} ($P_1$), {\it distance$_2$} ($P_2$). Here $W$ denotes the exact word appeared in the sentence. $P_1$ indicates distance (in terms of words) from the first drug name \cite{collobert11a,sunil16a}. This value will be zero for the first targeted drug name. $P_2$ is similar to $P_1$ but considers distance from the second targeted drug name. This way a word $w \in D^1 \times D^2 \times D^3$, where $D^i$ is the dictionary for $i^{th}$ local features. This feature layer constitutes the first layer for all models.

\subsection*{Embedding Layer}
In embedding layer, each discrete feature is mapped to a real-valued vector representation using a lookup or embedding matrix. 
Let us say $M^{i}$ is an embedding matrix for $i^{th}$ feature. Here each column of $M^{i}$ is a vector for the value in $i^{th}$ feature. Mapping can be done by taking product of one hot vector of feature value with its embedding matrix \cite{collobert11a}. Suppose $a^{(i)}_j$ is the one hot vector for $j^{th}$ feature value of $i^{th}$ feature then embedding layer will output: 

\begin{equation}
f^{(i)}_j = M^{i}.a^{(i)}_j
\end{equation}

\begin{equation}
x^i = f^{(i)}_1 \oplus f^{(i)}_2 \oplus f^{(i)}_3 
\end{equation}

Here $\oplus$ is concatenation operation so $x^i \in \mathbb{R}^{(n_1+....n_3)}$ is feature vector for $i^{th}$ word in sentence and $n_k$ is dimension of $k^{th}$ feature. Pre-trained word vectors are used for word embedding matrix and other feature matrices are initialized with random values. 
\subsection*{Bi-LSTM Layer}
Recurrent neural network is a powerful model for modeling sequential data \cite{Mikolov10}. It is a network with loop, allowing information to persist throughout the sequence. However it may suffer with vanishing or exploding gradient problems  \cite{Graves13,bengio2013} for instances of longer sentence. LSTM aims to overcome this problem by using gate and memory mechanism. LSTM layer is just another way to compute a hidden state which introduces a new structure called a memory cell ($c_t$) and three gates called as input ($i_t$), output ($o_t$) and forget ($f_t$) gates.  These gates are composed of {\it sigmoid} activation function and responsible for regulating information in memory cell. The final output of LSTM will be calculated on the basis of new states of memory cell. 

Consider $x^{1} x^{2} ..... x^{m}$ is the sequence of feature vectors of a sentence, where $m$ is the length of sentence and $x^t \in \mathbb{R}^d$ is a vector obtained by concatenating all feature vector of $t^{th}$ word. Let $h_l^{(t-1)}$ and $c^{(t-1)}_l$ be previous hidden and cell states of LSTM layer (LSTM$_l$)  respectively, then current hidden state ($h_l^{(t)}$), cell state $c^{(t)}_l$ and output of Bi-LSTM ($z^{(t)}$)  can be computed as
\begin{align*} 
&i^{(t)}_l = \sigma ( U^{(i)}_l x^{(t)} +  W^{(i)}_l h_l^{(t-1)} + b_l^i)\\
&f^{(t)}_l = \sigma (U^{(f)}_l x^{(t)} + W^{(f)}_l h_l^{(t-1)} + b_l^f)\\
&o^{(t)}_l = \sigma (U^{(o)}_l x^{(t)}  + W^{(o)}_l h_l^{(t-1)} + b_l^o)\\
&g^{(t)}_l = tanh(U_l^{(g)} x^{(t)} +  W^{(g)}_l h_l^{(t-1)} + b_l^{g}) \\
&c^{(t)}_l = c^{(t-1)}_l * f^{(t)}_l + g^{(t)}_l * i^{(t)}_l \\
&h^{(t)}_l = tanh(c^{(t)}_l) * o^{(t)}_l, 
\end{align*} 
where $\sigma $ is sigmoid activation function, $*$ is an element wise product, 
$U^{(i)}_l$, $U^{(f)}_l$, $U^{(o)}_l$, $U^{(g)}_l \in \mathbb{R}^{N \times d}$, 
$W^{(i)}_l$, $W^{(o)}_l$, $W^{(f)}_l$, $W^{(g)}_l \in \mathbb{R}^{N \times N}$, 
$b_l^i$, $b_l^f$, $b_l^o$, $b_l^g \in \mathbb{R}^{N}$,
$h^{(0)}_l$, $c^{(0)}_l \in \mathbb{R}^N$ are learning parameters for LSTM$_l$. 
Here $d$ is dimension of input feature vector and $N$ is hidden layer size. 
$h^{(t)}_l$ is output of LSTM$_l$ at time step $t$.
We compute $h_{r}^{(t)}$ in similar manner as $h_l^{(t)}$ by reversing the words of sentence. A separate LSTM$_r$ is used for this calculation. The final output for $t^{th}$ word by Bi-LSTM would be:
\begin{equation}
z^{(t)} = (h_{l}^{(t)} \oplus h_{r}^{(t)})
\end{equation}

\subsection*{Pooling Layer}
The objective of pooling layer is to get a fixed length features from a variable length word features. We experiment with two different kinds of pooling schemes:

\subsubsection*{(A) Max Pooling}
{\it Max} pooling takes one optimal over complete sequence. As Bi-LSTM accumulates information in both forward and backward directions, each node is assumed to have information of complete sentence. {\it Max} pooling takes the maximum over sentence assuming all important and relevant information are accumulated in that position. Let $z^1 z^2...z^m$ ($z^i \in \mathbb{R}^{N}$) be the sequence of vectors obtained after concatenating forward and backward LSTM output of each word then:
\begin{equation}
\label{max_pool}
z = \max_{1\leq i \leq(m)} [z^i],
\end{equation}
where $z \in \mathbb{R}^N$ is dimension wise $max$ of entire $z^i$'s.

\subsubsection*{(B) Attentive Pooling}
{\it Max} pooling may fail to perform well when important clues for the DDIs are present in different clauses or lie faraway in the sentence. Consider an example belonging to \textit{mechanism} class: \textit{Preliminary evidence suggests that cimetidine$_{Drug}$ inhibits mebendazole$_{Drug}$ metabolism and may result in an increase in plasma concentrations of mebendazole$_{Drug}$}. The first clue \textit{inhibits mebendazole metabolism} indicates interaction class likely to be \textit{effect} but the second clue \textit{increase in plasma concentration} makes it belong to the \textit{mechanism} class.
Taking one optimal over the complete sentence may incorrectly classify this instance. To overcome this issue we use {\it attentive} pooling which takes optimal based on weighted linear combination of feature vectors. Weights of the feature vectors are computed using attention mechanism which assign weights based on importance of that features \cite{cho14,yang2016,zhou2016}. The attention mechanism produces a vector $\alpha$ of size equal to length of sentence. The values in this vector are the weights we would assign to each word feature vectors. Weighted linear combination of Bi-LSTM outputs and attention weights are the output of attentive pooling layer. Let $Z \in \mathbb{R}^{N\times m}$ be the matrix of outputs obtained by Bi-LSTM then, output of attentive pooling would be:
\begin{align*}  
H &= tanh(Z) \\
\alpha &= Softmax(w^{aT} H)  \\
z &= \alpha Z^T, \numberthis \label{att_pool} 
\end{align*}
where 
$w^a \in \mathbb{R}^{N}$ is the learning parameter, $\alpha \in \mathbb{R}^{m}$ is attention weights and $z \in \mathbb{R}^N$ would be the output of attentive pooling layer. Important thing to notice here is that $\alpha$ would be different for every sentence, i.e., indicating relevant context words
may appear in distinct positions in different sentences.

\subsection*{Fully Connected and Softmax}
Output of pooling layer would be a fixed length vector. This vector undergoes non-linear transformation by applying {\it tanh} activation and which is then fed to fully connected neural layer. In fully connected layer we maintain the number of node equals to number of class. 
\begin{align*}
h^3 &= tanh(h^2) \\
p(y|x) &= Softmax(h^{3T} W^o + b^o) \numberthis \label{eqn} \\
\end{align*}
Here $h^2$ would be the output of pooling layer, $W^o \in \mathbb{R}^{N\times C}$, $b^o \in \mathbb{R}^{C}$ are parameters of fully connected neural network and $C$ is number of class in our model. To make classification, we use {\it softmax} function in the output of fully connected layer. {\it Softmax} will give normalized probability score for each class.

\subsection*{Training and Implementation}
All three models use cross entropy loss function for training the entire network. Adam's technique \cite{adam2014} is used for optimization.
We use batch size of $200$ for training each model. Implementation\footnote{Code for reprodocing the results is available at \url{https://github.com/sunilitggu/DDI-extraction-through-LSTM}} is done in python language using {\it Tensorflow}\footnote{\url{https://www.tensorflow.org}} package. 

\subsection{B-LSTM Model} 
{\it B-LSTM} is similar to the model proposed in \cite{liu2016,sunil16a} for DDI extraction and clinical relation extraction tasks. Here we use Bi-LSTM in place of convolution neural network used in \cite{liu2016,sunil16a}. As shown in figure \ref{fig:1a}, this model apply {\it Max} pooling on the output of Bi-LSTM to get optimal fixed length features. Max pooling is obtained from Equation \ref{max_pool} for every instance. These features are then fed to fully connected neural layer followed by {\it softmax} layer to obtain final classification.

\subsection{AB-LSTM Model}
Figure \ref{fig:1b} is the graphical representation of {\it AB-LSTM} model. In this case we apply {\it attentive} pooling on Bi-LSTM output. Attention weights are obtained from Equation \ref{att_pool} for each sentence. Output of attentive pooling layer was used as features and passed to fully connected and {\it softmax} layers to obtain final classification. 

\subsection{Joint AB-LSTM Model}
The objective of using {\it Joint AB-LSTM} is to take the advantage of both {\it max} and {attentive} pooling techniques. As shown in figure \ref{fig:1m} {\it Joint AB-LSTM} model uses two separate modules each with a Bi-LSTM network. Both Bi-LSTM take same feature vectors as input and produce output for every word in the sentence. We applied {\it Max} pooling on the first and {\it attentive} pooling on the second Bi-LSTM layer to get features from both the modules. Concatenation of both features are used for classification using fully connected and {\it softmax} layers. 

\section{Dataset Description}
\label{dataset}

We obtained the dataset from the shared challenge SemEval-2013 Task-9 \cite{segura2013,ddicorpus}. This dataset contains annotated sentences from two sources, Medline abstracts and DrugBank database. MedLine contains biomedical research articles and DrugBank contains manually entered texts collected from various sources and verified by accredited experts. The dataset is annotated with following four kinds of interactions:

{\bf Advice}: The text states an opinion or recommendation related to the simultaneous use of the two drugs, e.g.``{\it alpha-blockers} should not be combined with {\it uroxatral}".

{\bf Effect} : The sentence notes the effect of the drug-drug interaction or pharmacodynamic mechanism of interaction. For example ``{\it Warfarin} users who initiated {\it fluoxetine} had an increased risk of hospitalization for gastrointestinal bleeding".

{\bf Mechanism} : The sentence describes a pharmacokinetic mechanism, as in ``{\it Paroxetine} reduce the plasma concentration of {\it endoxifen} by about 20\%".

{\bf Int} : The text mentions a drug interaction without providing any other information. For example, ``This is typical of the interaction of
{\it meperidine} and {\it MAOIs.}".

Dataset provides the training and test instances as sentences. If a sentence has more than two drug names, all possible pairs of drugs in the sentence have been separately annotated. This way single sentence having multiple drug names leads to separate instances of drug pairs and corresponding interaction. Statistics of the dataset is shown in Table \ref{tab:tab1}. Finally the objective of model is to identify exact class of interaction (one of the four types) or no interaction. In our subsequent discussions, we refer the task as \textit{DDI classification}.

\begin{table}[htbp]
\centering
\scalebox{1}{
\begin{tabular}{|c|c|c|c|c|c|}
\hline
\multirow{2}{*}{\textbf{Corpus}} & \multicolumn{2}{c|}{\textbf{Training Set}} & \multicolumn{2}{c|}{\textbf{Test Set}} \\ \cline{2-5} 
 & \textbf{Before} & \textbf{After} & \textbf{Before} &  \textbf{After} \\ \hline
Documents		& 714 	& 714   & 191  & 191	\\ \hline
Pairs  			& 27774 & 16495 & 5716 & 4025	\\ \hline
Positive DDIs	& 4018	& 3844  & 979  & 979 	\\ \hline	
Negative DDIs	& 23756	& 12651 & 4737 & 3046	\\ \hline
Mechanism 		& 1318	& 1264  & 302  & 302	\\ \hline
Effect 			& 1685	& 1620  & 360  & 360	\\ \hline
Advice			& 826 	& 820   & 221  & 221	\\ \hline
Int				& 189 	& 140   & 96   & 96	\\ \hline
\end{tabular}
}
\caption{Statistics of DDI classification dataset before and after negative instance filtering}
\label{tab:tab1}
\end{table}

\subsection{Pre-processing} 
The following pre-processing is done in the dataset before using it in our model:
\begin{itemize}
 \item {\it Genia tagger}\footnote{\url{http://www.nactem.ac.uk/GENIA/tagger/}} is used for tokenization. All digits are normalized by replacing them with a special token $DG$ and all letters are changed to lowercase.

 \item The two targeted drug names are replaced with {\it DRUG-A} and {\it DRUG-B} respectively, and other drug names
   in the same sentence are replaced with {\it DRUG-N}. Earlier studies \cite{rastegar2013, liu2016} have reported that this step helps in improving the model performance.
\end{itemize}

\subsection{Negative Instance Filtering}
Consideration of all possible pairs of drug names in a sentence as separate instances for our model made the resultant dataset very imbalanced. 
We have 1:5.9 ratio of positive to negative instances. However, one can follow some strategies to remove negative
instances. Earlier studies \cite{zhao2016,kim2015,liu2016} have shown positive impact of negative instance filtering. 
We filter negative samples based on the following rules:

1. If both targeted drug mentions have the same name, remove the corresponding instance. Assumption behind this rule is drug doesn't interact with itself. We use string matching on both drug names to identify such cases. 

2. Remove the instance, if one drug is a kind of or a special case of the other one in the corresponding sentence. To identify such cases, we use regular expression by observing patterns in the dataset.  ``{\it DRUG-A (DRUG-B)}", ``{\it DRUG-A such as DRUG-B}" are examples of such patterns. 

3. If both target drugs appear in same coordinate structure, then remove the corresponding instance. We use several regular expressions based on observing the patterns in training set to filter out such instances. Examples of one such pattern is
``{\it DRUG-A , (DRUG-N , )$^{+}$DRUG-B}". 

Similar to \cite{liu2016}, our rules have not eliminated any positive instances from the test set. However, $144$ positive instances ($54$ {\it Mechanism}, $65$ {\it Effects}, $49$ {\it Int} and $6$ {\it Advice}) are removed from the training set. Table~\ref{tab:tab1} summarizes statistics of dataset before and after filtering.

\begin{table}[h]
\centering
\scalebox{1}{
\begin{tabular}{|c|c|c|}
\hline
\textit{\textbf{Models}} & \textit{\textbf{Dropout}} & \textit{\textbf{$l_2$ regu.}} \\ \hline
CNN*$^{1}$ 		& 0.7 & 0.1 \\ \cline{1-3}
B-LSTM 			& 0.7 & 0.001\\ \cline{1-3}
AB-LSTM 		& 0.7 & 0.0001 \\ \cline{1-3}
Joint AB-LSTM 	& 1.0 & 0.0001 \\ \cline{1-3}
\end{tabular}
}
\caption{Values of different regularization parameters used in the three models.}
\label{tab:reg_values} 
\end{table}

\begin{table*}[t]
\begin{minipage}{\textwidth}
\centering
\scalebox{1.0}{
\begin{tabular} 
{|p{0.15\linewidth}|p{0.07\linewidth}p{0.07\linewidth}p{0.09\linewidth}|p{0.07\linewidth}p{0.07\linewidth} p{0.09\linewidth}|p{0.05\linewidth}|} \hline
\multirow{2}{*}{\textbf{Models}} & \multicolumn{3}{c|}{\textbf{Before}} & \multicolumn{3}{c|}{\textbf{After}} & \multirow{2}{*}{$\triangle$} \\ \cline{2-7}
 & \textbf{Precision} & \textbf{Recall} & \textbf{F1 Score} & \textbf{Precision} & \textbf{Recall} & \textbf{F1 Score} & \\ \hline
{\it \textbf{CNN*}} 			& 62.11 & 63.63 & 62.86	& 66.14 & 60.87 & 63.40 & 0.86\\ 
{\it \textbf{B-LSTM}} 			& 69.07 & 64.35 & 66.63 & 70.62 & 66.80 & 68.66 & 3.05\\
{\it \textbf{AB-LSTM}}			& 70.75 & 60.06 & 64.97 & 73.34 & 62.41 & 67.43 & 3.79\\
{\it \textbf{Joint AB-LSTM}}	& 67.77 & 66.80 & 67.28 & 74.47 & 64.96 & 69.39 & 3.14\\
\hline 
\end{tabular}
}
\caption{Performance improvement after filtering negative instance from the dataset. $\triangle$ indicates percentage of relative improvement in F1 score.}
\label{tab:ddi_res_neg_filtering}
\end{minipage}
\end{table*}

\begin{table*}[t]
\begin{minipage}{\textwidth}
\centering
\scalebox{1.0}{
\begin{tabular} 
{|p{0.12\linewidth}|p{0.04\linewidth}p{0.04\linewidth}p{0.04\linewidth}|p{0.04\linewidth}p{0.04\linewidth}p{0.04\linewidth}|p{0.04\linewidth}p{0.04\linewidth}p{0.04\linewidth}|} \hline

\multirow{2}{*}{\textbf{Class}} & \multicolumn{3}{c|}{\textbf{B-LSTM}} & \multicolumn{3}{c|}{\textbf{AB-LSTM}} & \multicolumn{3}{c|}{\textbf{Joint AB-LSTM}} \\ \cline{2-10}
 & \textbf{P} & \textbf{R} & \textbf{F} & \textbf{P} & \textbf{R} & \textbf{F} & \textbf{P} & \textbf{R} & \textbf{F} \\ \hline
{\it \textbf{Advice}}		& $\downarrow$ & $\uparrow$ & $\uparrow$ & $\downarrow$ & $\uparrow$ & $\downarrow$ & $\downarrow$ & $\uparrow$ & $\uparrow$ \\
{\it \textbf{Mechanism}}	& $\uparrow$ & $\downarrow$ & $\uparrow$ & $\uparrow$ & $\sim$ & $\uparrow$ & $\uparrow$ & $\downarrow$ & $\uparrow$ \\
{\it \textbf{Effect}}		& $\sim$ & $\uparrow$ & $\uparrow$ & $\uparrow$ & $\uparrow$ & $\uparrow$ & $\uparrow$ & $\downarrow$ &$\uparrow$ \\
{\it \textbf{Int}}			& $\uparrow$ & $\uparrow$ & $\uparrow$ & $\uparrow$ & $\uparrow$ & $\uparrow$ & $\downarrow$ & $\sim$ & $\downarrow$ \\
\hline 
\end{tabular}
}
\caption{Relative change in class-wise performance measures after negative filtering. $\downarrow$ indicates value got reduced after filtering step compared to the value obtained using the complete data. Similarly $\uparrow$ indicates increase in value and $\sim$ indicates value remain almost same after filtering step.}
\label{tab:all_conf_mat}
\end{minipage}
\end{table*}

\section{Experiment Design}
We train and evaluate three LSTM models on the DDI classification task. As mentioned earlier, if a sentence contains more than two drug names then all possible pairs with the sentence
constitute separate instances/samples. We use the same evaluation scheme as used in the challenge \cite{segura2013}.

\subsection{Hyper-parameters}
As there is no separate development or validation set available, we divided the original training dataset into two parts, $80\%$ as training and rest $20\%$ as validation sets. Hyper-parameters are tuned using this validation set. Hidden layer size in {\it B-LSTM} and {\it AB-LSTM} are kept as $200$, and $150$ for \textit{Joint AB-LSTM}. Pre-trained word embedding of $100$ dimensions and distance embedding of $10$ dimensions are used in all three LSTM models. Word embedding are obtained using {\it GloVe} tool \cite{pennington14} on a corpus of PubMed open source articles \cite{muneeb15}. We used both $l_2$ regularization and dropout \cite{srivastava2014} techniques for regularization. We applied dropout only on the output of the pooling layers. Different values of the regularization parameters are shown in the Table~\ref{tab:reg_values}. Next to decide about the early-stopping criteria (i.e. number of epochs neural network models will be trained), we again split the original training dataset into two parts containing $95\%$ and $5\%$ of all samples. Now we trained the model on $95\%$ set using above selected hyper-parameters and selected the epoch giving the best result on the left out $5\%$ of the original training data. Trained model on the $95\%$ set is used on the independent test set and corresponding results are discussed in the respective sections.

\subsection{Baseline Methods for comparison}
\label{sec:baseline}
We compare performance of the three proposed models with several baseline methods. Approaches based on conventional features, kernel methods and on neural networks are included as baseline methods. Below we briefly describe about the baseline methods, where superscript one ($*^1$) indicates one stage and superscript two $(*^{2})$ indicates two stages methods: 

{\bf Linear Methods:}
In this class of methods, a linear classifier is used to identify the correct class of interaction for each instance. All instances are represented by a vector of manually designed features. {\bf UTurku}$^{1}$ used Turku event extraction system (TEES) \cite{bjorne2013} for drug interaction extraction. The major features used by TEES comes from dependency parsing and domain dependent resources such as MetaMap. {\bf UWM-TRIADS}$^{2}$ \cite{rastegar2013} and {\bf Kim}$^{2}$ \cite{kim2015} are two stage methods. In both the stages SVM with contextual, lexical, semantic and tree structured features was used.  

{\bf Kernel Methods:}
Kernel methods are powerful techniques for utilizing graph based features in any natural language processing task. {\bf WBI-DDI}$^{2}$ and {\bf FBK irst}$^{2}$ are two stage methods \cite{chowdhury2013a,thomas2013} for DDI classification. First stage of both models used different kernels methods for utilizing syntax tree and dependency tree features. In the second stage WBI-DDI$^2$ used TEES and FBK irst$^2$ used SVM with non-linear kernel for classification. {\bf NIL UCM}$^{1}$ used multi-class SVM as kernel methods in one stage framework.

{\bf Neural Network Methods:}
Neural network or deep learning methods use latent features in place of manually designed features. This class of algorithms use neural network to encode word level features for generation of sentence level features. Final classification happens with sentence level features. {\bf SCNN$^{1,2}$}~\cite{zhao2016} used convolution neural network with max pooling layer to learn higher level discriminative features over entire sentence. SCNN also utilized PoS tags and dependency tree based features apart from latent word embedding and distance embedding features. {\bf MV-RNN$^{1}$} \cite{suarez2016} is also a neural network based model. In particular, MV-RNN used recursive neural network \cite{socher11} for learning embedding of sentence or part of sentence recursively and thus obtained final vector is used for classification. To analyze the performance of RNN and CNN based models, we implemented CNN with max pooling technique for this task. We implemented the similar model, as discussed in \cite{liu2016}, due to unavailability of the source code and referred to it as CNN*$^{1}$. Initially, the results were obtained with similar parameter settings as discussed in \cite{liu2016} but we could not obtain the similar results, as mentioned in that research article. So we tuned the model hyper-parameters and used them in subsequent analysis. The difference in performance could be due to change in training dataset after negative instance filtering as well as due to use of different pre-trained embedding vectors.

\section{Results and Discussions}
\label{results}

\subsection{Effect of Negative Instance Filtering}
\label{neg_effect}
Following our filtering rules, a large number of negative instances along with few positive instances are removed from the dataset. However none of the positive instances from {\it test set} are removed. There was a significant change in imbalance ratios of interaction classes due to the filtering step. For example, imbalance ratio changed from 1:32.6 to 1:19.1 for the interaction class {\it Advice} and from 1:146 to 1:116.8 for the interaction class {\it Int}. Table \ref{tab:ddi_res_neg_filtering} shows performance of different models on {\it test set} while training is done using either complete or filtered dataset. All models when using filtered dataset gave improved performance in terms of F1 score. All three LSTM models obtained more than $3\%$ of relative improvement with the use of filtered dataset. Performance improvement can be attributed to two factors: cleaner data and reduction of imbalance. Both these factors are result of the data filtering step.

We next look at class-wise performance of all LSTM models to see how change in imbalance ratio affected the model performance at individual class level. For this we compared the relative changes in \textit{precision}, \textit{recall} and \textit{F1 score}, when models trained on the filtered dataset vs the complete dataset. The results are summarized in the Table~\ref{tab:all_conf_mat}. The overall F1 score for individual class increases in almost all the cases. For \textit{Advice} class, all three models have similar effect on precision and recall. All three LSTM models have obtained better precision but recall got reduced or remained almost same for the \textit{Mechanism} class. For the \textit{Effect} class, \textit{Joint AB-LSTM} model obtained reduced recall, but all the models have obtained better or similar precision values after filtering. The performance pattern of the \textit{B-LSTM} and \textit{AB-LSTM} models were same for the \textit{Int} class. 
It is very difficult to reason out the effect of filtering step due to inconsistent effect of it on the three models.

\begin{table}[ht]
\centering
\scalebox{1}{
\begin{tabular}{|c|c|c|c|}
\hline
\textit{\textbf{Models}} & \textit{\textbf{Precision}} & \textit{\textbf{Recall}} & \textit{\textbf{F1 Score}} \\ \hline
SCNN$^{2}$ \cite{zhao2016}&  68.5 & 61.0 & 64.5\\ \hline 
CNN*$^{1}$ 			& 62.11 & 63.63 & 62.86	  \\ \hline
B-LSTM$^{1}$ 		& 69.07 & 64.35 & 66.63  \\ \hline 
AB-LSTM$^{1}$ 		& 70.75 & 60.06 & 64.97  \\ \hline 
Joint AB-LSTM$^{1}$ & 67.77 & 66.80 & 67.28  \\ \hline 
\end{tabular}
}
\caption{Performance comparison of proposed models with existing models on complete dataset, i.e. without using negative instance filtering, for DDI classification task.}
\label{tab:ddi_pre_filtered_results} 
\end{table}

\begin{table}[hbt]
\centering
\scalebox{1}{
\begin{tabular}{|c|c|c|c|}
\hline
 \textbf{Method} & \textbf{Precision} & \textbf{Recall} & \textbf{F Score} \\ \hline  
 UTurku$^{1}$ \cite{bjorne2013}&  73.2 & 49.9 & 59.4 \\ 
 UWM-TRIADS$^{2}$ \cite{rastegar2013}& 43.9  & 50.5  & 47.0 \\ 
 Kim$^{2}$ \cite{kim2015}  & -	&  	-	& 67.0 \\ \hline

NIL UCM$^{1}$ \cite{bokharaeian2013}& 53.5 & 50.1 & 51.7 \\ 
 WBI-DDI$^{2}$ \cite{thomas2013}	&  64.2 & 57.9 & 60.9\\ 
 FBK irst$^{2}$ \cite{chowdhury2013}& 64.6 & 65.6 & 65.1\\ \hline 
 
 SCNN$^{1}$ \cite{zhao2016} & 69.1 & 65.1 & 67.0  \\ 
 SCNN$^{2}$ \cite{zhao2016} & 72.5 & 65.1 &  68.6  \\ 
 MV-RNN$^{1}$ \cite{suarez2016} & 52.0 & 48.0 & 50.0 \\  
\hline \hline
{\bf CNN$^{\star}$}$^{1}$ 	& 66.14 & 60.87 & 63.40  \\   
 {\bf B-LSTM}$^{1}$ 		& 70.62 & {\bf 66.80} & 68.66   \\   
 {\bf AB-LSTM}$^{1}$		& 73.34 & 62.41 & 67.43  \\   
 {\bf Joint AB-LSTM}$^{1}$ & {\bf 74.47} & 64.96 & {\bf 69.39} 	\\ \hline 
\end{tabular}
}
\caption{Performance comparison between the proposed methods and top-ranking approaches on the test data. Performance is measured based on precision, recall and F1 score.  The highest scores are highlighted in bold.}
\label{tab:ddi_res}
 \end{table}

\subsection{Comparison with Baseline Methods}
\label{comp_res}
 {\bf Comparison based on the complete dataset:} 
 First we compare the three proposed models with other existing models on the complete dataset, i.e., without using the filtering step. Best results out of five runs are shown in Table~\ref{tab:ddi_pre_filtered_results}. Here, we have included only those methods which explicitly mention about the similar steps and report corresponding  results. {\it B-LSTM} and {\it Joint AB-LSTM} models have outperformed all other models. We performed t-test (one-sided null hypothesis: mean performance of model 1 is not less than that of model 2) on mean performance of model pairs. Mean performance of a model was calculated as the average of F1 score obtained in five different runs of the model. \textit{B-LSTM} and \textit{Joint AB-LSTM} models outperformed CNN*$^{1}$ model at the significance level of $0.05$. In particular p-values of CNN*$^{1}$ vs {\it B-LSTM}, {\it AB-LSTM}, and {\it Joint AB-LSTM} comparisons were $0.0007, 0.04, 0.01$ respectively. Among the LSTM models, no single model outperforms other models at the above chosen significance level.

{\bf Comparison based on the filtered dataset:}
Table \ref{tab:ddi_res} provides a comparison of our models with previous approaches. 
{\it Joint AB-LSTM} model obtained the best F1 score of $69.39\%$. There is $3.6\%$ of relative improvement in F1 score on comparison with the best performing method (Kim$^2$) among feature based linear and kernel methods. In comparison to SCNN$^{2}$ model, all three proposed models gave similar results for this task. SCNN$^{2}$, a convolution neural network based model, uses higher order grammatical features based on parts of speech and shortest dependency path apart from pre-trained word embedding and position embedding features. If we remove shortest dependency path features from SCNN$^{2}$ model, its performance decreases to $63.8\%$ from $68.6\%$. On the other hand, apart from {\it Joint AB-LSTM}, the other two models also gave relatively much better performance than SCNN$^{2}$ with only using simple features.

Among the three proposed models, {\it Joint AB-LSTM} always performs better than the other two methods. Based on the McNemar test {\it B-LSTM}, {\it AB-LSTM}, and {\it Joint AB-LSTM} models outperformed CNN*$^{1}$ model with p-values of $0.0005$, $0.001$ and $8.4\times10^{-9}$ respectively. As results indicate all LSTM based models outperform CNN*$^{1}$ model significantly. Among the LSTM models, {\it Joint AB-LSTM} and \textit{B-LSTM} models outperform {\it AB-LSTM} with p-values of $0.005$ and $0.03$ respectively on the DDI classification task. Further, the McNemar test suggests that there is no significant difference in performance of the two models on the DDI-classification task.

\begin{table}[hbt] 
\centering
\scalebox{0.8}{
\begin{tabular}{|c|c|c|c|c|c|}
\hline
\textit{\textbf{Models}} & \textit{\textbf{Advice}} & \textit{\textbf{Mechanism}} & \textit{\textbf{Effect}} & \textit{\textbf{Int}} &\textit{\textbf{MAVG}} \\ \hline
\hline
UTurku \cite{bjorne2013} & 63.0 & 58.2 & 60.0 & 50.7 & 58.7 \\ \hline 
UWM-TRIADS\cite{rastegar2013}& 53.2 & 44.6 & 44.9 & 42.1 & 47.2 \\ \hline
Kim \cite{kim2015}  & 72.5 & 69.3 & 66.2 & 48.3 & 64.1 \\ \hline
\hline
FBK irst \cite{chowdhury2013}& 69.2 & 67.9 & 62.8 & {\bf 54.7} & 64.8\\ \hline
NIL\_UCM \cite{bokharaeian2013} & 61.3 & 51.5 & 48.9 & 42.7 & 53.5\\ \hline
WBI-DDI\cite{thomas2013}	 & 63.2 & 61.8 & 61.1 & 51.1 & 59.7\\ \hline
\hline
MV-RNN$^{1}$ \cite{suarez2016} & 57.0&46.0 &49.0 & 49.0& 50.25\\ 
\hline
{\bf CNN*$^{1}$} 	&  69.04 & 63.98 & 63.00 & 45.07 &  60.27 \\ \hline 
{\bf B-LSTM} 		&  75.92 & {\bf 72.66} & 65.15 & 47.40 &  65.28 \\ \hline 
{\bf AB-LSTM} 		&  69.68 & 68.06 & {\bf 68.28} & 54.16 & 65.04  \\ \hline 
{\bf Joint AB-LSTM} &  {\bf 80.26} & 72.26 & 65.46 & 44.11 & {\bf 65.52} \\ \hline 
\end{tabular}
}
\caption{Performance comparison between the proposed methods and top-ranking approaches on the test data for DDI classification. Performance are measured through F1-Score for each class and Macro Average (MAVG).  The highest scores are highlighted in bold.}
\label{tab:ddi_class_filtered_results} 
\end{table}

\begin{table}
\centering
\scalebox{0.8}{
\begin{tabular}{|c|c|c|c|}
\hline
\textit{\textbf{Models}} & \textit{\textbf{Precision}} & \textit{\textbf{Recall}} & \textit{\textbf{F Score}} \\ \hline
\hline
{\bf Joint AB-LSTM}							& 74.47 & 64.96 & 69.39   	\\  \hline
{\bf Joint AB-LSTM - \{ P \} } 		 		& 70.62 & 66.80 & 68.66 	\\  \hline
{\bf Joint AB-LSTM - \{ (P + X) \} }			& 71.21 & 61.89 & 66.22	\\ 
\hline
\end{tabular}
}
\caption{Contribution of each feature in {\bf Joint AB-LSTM} model. Here $P$ refers random position embedding for both $P_1$ and $P_2$, and $X$ refers pre-trained word vector embedding}
\label{tab:each_features}
\end{table}

\begin{figure*}[tb]
\centering
  \begin{subfigure}[b]{.5\linewidth}
    \centering
    \includegraphics[width=0.9\textwidth]{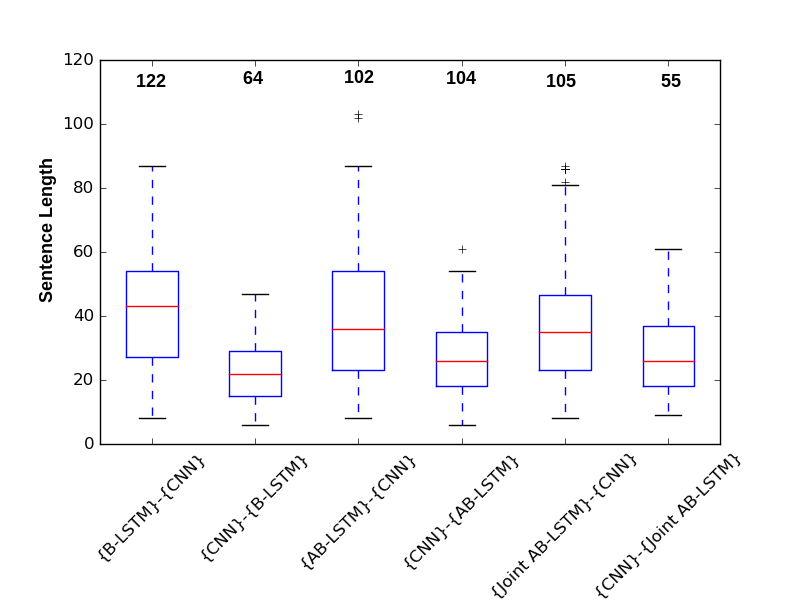}
    \caption{Box plot for sentence length (number of words)}
    \label{fig:sentence_length}
  \end{subfigure}%
  \begin{subfigure}[b]{.5\linewidth}
    \centering
    \includegraphics[width=0.9\textwidth]{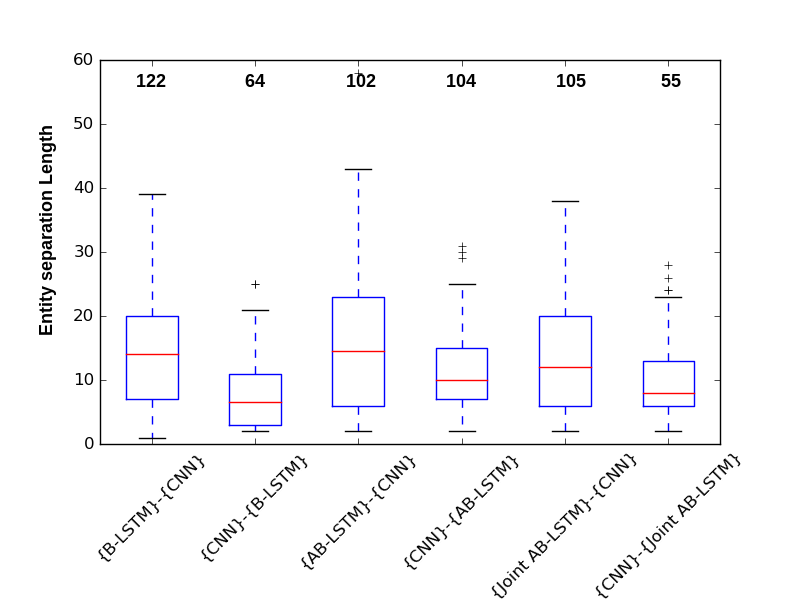}
    \caption{Box plot for entity separation (number of words)}
    \label{fig:entity_sep}
  \end{subfigure}%
 \caption{Boxplot for sentence length and entity separation length in instances. Here \{X\}-\{Y\} represent instances  correctly predicted by X but not by Y.}
 \label{fig:box_length}
\end{figure*}

{\bf Class wise Performance Analysis:} We compare class wise performance of the proposed models with existing models in Table \ref{tab:ddi_class_filtered_results}. 
We observe that no single model outperforms others for all classes. {\it FBK irst} obtained the best performance for {\it Int} class. \textit{B-LSTM}, {\it AB-LSTM} and \textit{Joint AB-LSTM} obtained the best performance for \textit{Mechanism}, {\it Effect} and  {\it Advice} classes respectively. However, {\it Joint AB-LSTM} outperformed all other models on aggregate performance measure using macro-average F1 score. All models finds it easier to detect {\it Advice} interaction types compared to the instances of the other three interaction types. Similarly all models find it most difficult to detect {\it Int} interaction types. The worse performance on the {\it Int} class can be attributed to insufficient training data.
{\it Effect} interaction class was found to be the second most difficult class to detect by most of the models compared in this analysis.

\subsection{Feature Analysis}
In order to validate the importance of each feature, we further analyzed the performance of {\it Joint AB-LSTM} model by removing feature types one by one. It can be observed from Table \ref{tab:each_features} that the use of pre-trained word embedding is important feature. About $1.1\%$ of relative decrement is observed if the model does not use position embedding. On the other hand, removal of position embedding as well as use of random vectors in place of pre-trained word vectors lead to $4.6\%$ of relative decrements in the model's performance. This analysis clearly indicates the importance of word and position embedding features.

\subsection{LSTM vs CNN models}
Earlier we discussed that intuitively it seems LSTM models are likely to perform better for longer sentences compared to CNN model. We performed an analysis on our results to see whether that is the case or not. For this we analyze length as well as entity separation between two targeted drugs of all those sentences which were predicted correctly by one model but incorrectly by the other. Figures \ref{fig:sentence_length} and \ref{fig:entity_sep} show the box plots for sentence length and separation length between targeted drugs. Here \{X\}-\{Y\} represent instances correctly predicted by X but not by Y. In all of these cases length represent number of words and numbers present at top of the boxes represent number of instances present in that category. From the figures we can observe that the proposed LSTM models performed better than CNN in both scenarios of instances having longer sentences and having larger entity separation length.
\begin{figure*}[t]
\centering
\includegraphics[width=18cm, height=3cm]{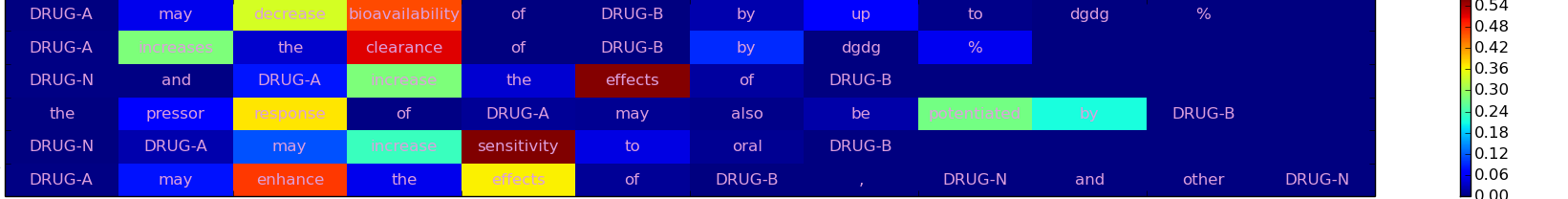}
\caption{Heat map of attention weights indicating importance of relevant words.}
\label{fig:att_weight}
\end{figure*}

\subsection{Error Analysis}
Apart from imbalance issue, we try to find out whether there is any other factor which is adversely affecting performance of the models. For this we look at average sentence lengths of correctly and incorrectly classified instances by each of the four models for the DDI classification task (Table~\ref{tab:error_ana}). We observe that, average sentence length and entity separation length for incorrectly classified instances are always high compared to the correctly classified sentences. Another aspect of incorrectly predicted instances was, presence of multiple drug entities in many such instances. Repetitive drug entities is more likely to behave like a noise, which may cause neural models to lose relevant information from other words likely to be contextually important. Hence a better strategy is required to deal with such cases. Considering limited context along with removal of repetitive entities can be one way to deal with extra large sentences. 

\begin{table}[ht]
\centering
\scalebox{0.8}{
\begin{tabular}{|c|c|c|c|c|}
\hline
\multirow{2}{*}{\textbf{Model}} & \multicolumn{2}{|c|}{\textbf{Sentence Length}} & \multicolumn{2}{|c|}{\textbf{Entity Separation}} \\ \cline{2-5} 
& True & False & True & False \\ \hline
CNN*	   & $26.19_{(13.38)}$ & $42.51_{(21.00)}$ & $11.24_{(9.39)}$ & $15.17_{(12.21)}$ \\ \hline
B-LSTM  & $29.12_{(16.19)}$ & $37.34_{(22.49)}$ & $12.15_{(9.55)}$ & $13.92_{(13.03)}$ \\ \hline  
AB-LSTM  & $28.52_{(17.57)}$ & $37.54_{(20.19)}$ & $11.12_{(10.11)}$ & $14.26_{(12.65)}$ \\ \hline 
Joint AB-LSTM &  $28.52_{(14.50)}$ & $39.97_{(21.70)}$ & $12.81_{(9.49)}$ & $14.19_{(11.86)}$ \\ \hline 
 \end{tabular}
 }
\caption{Mean and standard deviations (in subscript) of length of sentence for True Positive and False Negative instance in DDI classification task}
\label{tab:error_ana}
\end{table}

\subsection{Visual Analysis}
In order to confirm that the model is able to learn attention weights based on importance of words, we visualize attention weights of some of the sentences after training {\it Joint AB-LSTM}. Figure \ref{fig:att_weight} is the heat map of attention weights for $6$ instances of test set. Here every line is a sentence with two targeted drug names replaced with special tokens {\it DRUG-A} and {\it DRUG-B} and darkness in red color indicate heedfulness. Figure shows that our model can select important words based on the task. For example in the sentence {\it ``DRUG-A may enhance the effects of DRUG-B , DRUG-N and other DRUG-N"}, model is able to assign high weights to {\it ``may enhance the effects"} very well. Similarly, in the sentence {\it ``DRUG-N and DRUG-A increase the effects of DRUG-B"}, the model is assigning high weights to the words {\it increase} and {\it effect}.  

\section{Conclusion}
\label{conc}
In this work we proposed three LSTM based models {\it B-LSTM}, {\it AB-LSTM} and {\it Joint AB-LSTM} for DDI classification task. All the three models use simple word and distance embedding as features and learn higher level feature representation using Bi-LSTM network. Two of the proposed models also utilized neural attention mechanism to get higher level feature representation. To the best of our knowledge, it is the first study to use LSTM and attention mechanism for DDI extraction task. Performance of all three models are compared with the existing methods on SemEval-2013 DDI extraction dataset. {\it Joint AB-LSTM} model achieves state-of-the-art for the DDI classification task. Performance of the other two models, {\it B-LSTM} and {\it AB-LSTM}, are also found to be competitive. Analysis of the results indicates the following important points: imbalance and noise adversely affect all models, {\it Advice} interaction class is easiest to predict, repetitive entries of other drug names negatively effect all models, and models are likely to make incorrect classification for longer sentences.

\bibliographystyle{abbrv}
\bibliography{sigproc}   
\end{document}